\pdfoutput=1

\documentclass[11pt]{article}

\usepackage{emnlp2021}

\usepackage{times}
\usepackage{latexsym}

\usepackage[T1]{fontenc}
\usepackage{amsfonts}
\usepackage{multirow}
\usepackage{multicol}
\usepackage{changepage}

\usepackage[utf8]{inputenc}

\usepackage{microtype}

\usepackage{graphicx}
\usepackage{amsmath}
\usepackage[ruled,vlined]{algorithm2e}
\usepackage{tabularx}

\title{Cross-lingual Adaption Model-Agnostic Meta-Learning for Natural Language Understanding}

\author{Qianying Liu \hspace{10 mm} Fei Cheng \hspace{10 mm} Sadao Kurohashi \\
        Graduate School of Informatics, Kyoto University \\
        \tt ying@nlp.ist.i.kyoto-u.ac.jp, \{feicheng,kuro\}@i.kyoto-u.ac.jp;\\}

\begin{document}
\maketitle
\begin{abstract}

Meta learning with auxiliary languages has demonstrated promising improvements for cross-lingual natural language processing. However, previous studies sample the meta-training and meta-testing data from the same language, which limits the ability of the model for cross-lingual transfer. In this paper, we propose XLA-MAML, which performs direct cross-lingual adaption in the meta-learning stage. We conduct zero-shot and few-shot experiments on Natural Language Inference and Question Answering. The experimental results demonstrate the effectiveness of our method across different languages, tasks, and pretrained models. We also give analysis on various cross-lingual specific settings for meta-learning including sampling strategy and parallelism.

\end{abstract}

\section{Introduction}

There are around 7,000 languages spoken around the world, while most natural language processing (NLP) studies only consider English with large-scale training data. Such setting limits the ability of these NLP techniques when sufficient labeled data is lacking for effective finetuning. To address this problem and extend the global reach of NLP, recent research of Cross-lingual Natural Language Understanding (XL-NLU) focuses on multi-lingual pretrained representations such as multi-lingual BERT (mBERT), which cover more than 100 languages. These cross-lingual pretrained models allow zero-shot or few-shot transfer among languages for specific tasks, where a model for a task is trained in monolingual English data and directly applied to a target low resource language. 
With the support of strong multi-lingual representations, the further transfer among languages of XL-NLU could be formalized similar to the few-shot learning task.


Meta learning, or learning to learn, aims to create models that can learn new skills or adapt to new tasks rapidly from few training examples, which has demonstrated promising result for few-shot learning. Model-Agnostic Meta-Learning (MAML)~\cite{finn2017model} has been recently shown beneficial for various few-shot learning and cross-domain learning NLP tasks. The algorithm allows fast adaptation to a new task using only a few data points. 

\begin{figure}[!t]
\centering 
    \includegraphics[width=0.3\textwidth]{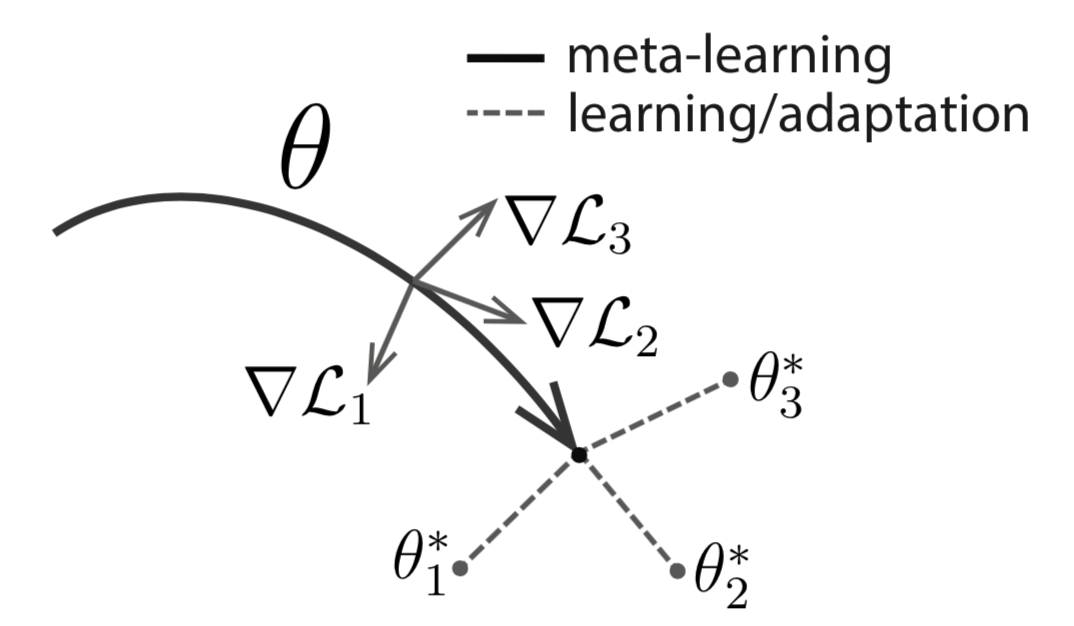}
    \caption{Illustration of cross-lingual meta learning.}
    \label{XLA-MAML}
\end{figure}

From MAML perspective, we can re-formulate cross-lingual transfer task as shown in Figure 1. For model $f_\theta$ with parameters $\theta$, we can consider the task distribution $\mathcal{P}(\tau)$ formed by drawing from different auxiliary language \textit{bags} of data samples $\{(x,y)_{l}\}$.
For meta training, we simulate the cross-lingual transfer process and sample a meta dataset
$\{\mathcal{S}, \mathcal{Q}\}$
from different bags $\{(x,y)_{l}\}$.
The support set $\mathcal{S}$ simulates the transfer source training data and updates the model parameters to $\theta'_i$, which is called the \textit{inner step}. Then the query set
$\mathcal{Q}$
simulates the transfer target testing data, model parameters are trained by optimizing for the performance of $f(\theta')$  with respect to $\theta$ across tasks sampled from the target bags $\{(x,y)_{l}\}$, which is referred as the \textit{meta step}. 
The algorithm minimize the prediction error on target language given a small support set for learning.


\citet{nooralahzadeh2020zero} proposed a meta-learning based cross-lingual transfer model, which uses data of non-target low-resource languages as auxiliary data. However, they sample the inner tasks and meta-update tasks from the same auxiliary language, that the meta learning cannot perform transfer for cross-lingual adaption. In this paper, to address this problem, we propose Cross-lingual Adaption Model-Agnostic Meta-Learning (XLA-MAML), which performs cross-lingual adaption in the meta learning stage to transfer knowledge across different languages.  

In conclusion, our contribution is 3-fold:

\begin{itemize}
\item We propose a cross-lingual adaption meta-learning approach, which directly performs meta learning adaption across different languages, while previous studies sample the support and query set from the same language.

\item We conduct zero-shot and few-shot experiments on two novel benchmarks of XL-NLU, XNLI and MLQA. Experimental results demonstrate the effectiveness of our model compared with previous MAML studies, especially under zero-shot settings.
\item We analyze the XLA-MAML performance under a number of cross-lingual meta learning settings, including sampling strategy and parallelism. The additional results on XLM-R suggest that XLA-MAML benefits even with a stronger encoder. 
\end{itemize}

\begin{algorithm*}[!h]
\small
\SetAlgoLined
\KwIn{high-resource language dataset $\mathbb{D}_{src}$, \\set of low-resource language auxiliary datasets $\{\mathbb{D}_{aux}$\} with languages $\mathbb{L}$}
\KwResult{Predictions $\{y'_{tgt}\} = \{f_\theta(x_{tgt})\}$ for target language test set $\{x_{tgt}\}$}
 Initialization model $f(\theta)$ with pretrained mMLM model\;
 Finetune $f(\theta)$ with high-resource language dataset $\mathbb{D}_{src}$\;
 \While{not stop criterion}{
  sample $\{l_i\}_S \subset \mathbb{L}, \{l_i\}_Q \subset \mathbb{L}$\;
  sample $\mathcal{S} = \{(x^k,y^k)\}_{k=1}^K \subset \bigcup_{\{l_i\}_S}\mathbb{D}_{l_i} \subset \mathbb{D}$, $\mathcal{Q} = \{(x^n,y^n)\}_{n=1}^N \subset \bigcup_{\{l_i\}_Q}\mathbb{D}_{l_i} \subset \mathbb{D}$\;
  $\tau_i: \mathcal{D} = (\mathcal{S}, \mathcal{Q})$\;
  Compute $\nabla_\theta\mathcal{L}^{(0)}_{\tau_i}(f_\theta)$  on $\mathcal{S}$\;
  Compute adapted parameters with gradient descent: $\theta'_i = \theta - \alpha \nabla_\theta\mathcal{L}^{(0)}_{\tau_i}(f_\theta)$\;
compute $\nabla_{\theta} \sum_{\tau_i \sim p(\tau)} \mathcal{L}_{\tau_i}^{(1)} (f_{\theta'})$ on $\mathcal{Q}$\;
Compute meta update for parameters: $    \theta \leftarrow \theta - \beta \nabla_{\theta} \sum_{\tau_i \sim p(\tau)} \mathcal{L}_{\tau_i}^{(1)} (f_{\theta'_i})$\;
  
 }
 Perform zero-shot or few-shot evaluation on target language test set $\{x_{tgt}\}$\;
 \caption{Cross-lingual Adaption Model-Agnostic Meta-Learning}
 \label{alg:XLA-MAML}
\end{algorithm*}
\section{Meta Learning}

Meta-learning algorithms train over a variety of learning tasks $\tau$ including potentially unseen tasks on a distribution of tasks $\mathcal{P}(\tau)$ and optimized for the best performance of a \textit{base-learner} model $f_\theta$ with parameters $\theta$. Each task $\tau$ is associated with a dataset sample$\mathcal{D}$, containing both feature vectors $\{x\}$ and labels $\{y\}$. The optimal model parameters are:

\begin{equation}
    \theta^* = \arg\min_\theta \mathbb{E}_{\mathcal{D}\sim p(\mathcal{D})} [\mathcal{L}_\theta(\mathcal{D})]
    \label{eq:1}
\end{equation}

Where one dataset $\mathcal{D}$ is considered as one data sample. The dataset sample $\mathcal{D}$ consists of a support set $\mathcal{S}$ and a query set $\mathcal{Q}$, where the support set $\mathcal{S}$ is used for \textit{meta-training} which simulates fast adaptation and the query set $\mathcal{Q}$ is used for \textit{meta-testing} that evaluates performance and compute a loss with respect to model parameter initialization. 
A loop of \textit{meta-training} to update the parameter initialization  and \textit{meta-testing} to update the loss is called an \textit{episode}. 
Meta learning consists of updating the parameters of the base-learner $f_\theta$ by episode loops until it reaches stop criterion. The optimal model parameters in Equation \ref{eq:1} could be rewritten as:
\begin{equation}
        \theta = 
        \arg\max_\theta E_{S,Q \subset\mathcal{D}} [\sum_{(x, y)\in Q} P_\theta(x, y{, S})]
        \label{eq:2}
\end{equation}



There are three common approaches of meta-learning: metric-based, model-based, and optimization-based. Each approach uses different ways to model $P_\theta(y|x)$. Metric based meta learning methods predict probability over a set of known labels $y$ is a weighted sum of labels of support set samples. The weight is generated by a kernel function $k_\theta$, measuring the similarity between two data samples, by modeling $P_\theta(y \vert \mathbf{x}, S) = \sum_{(\mathbf{x}_i, y_i) \in S} k_\theta(\mathbf{x}, \mathbf{x}_i)y_i$ \cite{koch2015siamese, sung2018learning,vinyals2016matching, snell2017prototypical}.
Model based meta learning methods depend on a model designed specifically for fast learning, which updates its parameters rapidly with a few training steps. This rapid parameter update can be achieved by its internal architecture or controlled by another meta-learner model. Memory-Augmented Neural Networks~\cite{santoro2016meta} uses external memory storage to facilitate the learning process of neural networks. Meta Network~\cite{munkhdalai2017meta} uses fast weights, which predicts the parameters of another neural network and the generated weights. 

In this work we focus on optimisation-based method MAML that optimisation-based methods are compatible with any model that learns through gradient descent.
Optimized based adjust the optimization algorithm so that the model can perform fast adaption \cite{DBLP:conf/iclr/RaviL17, finn2017model, DBLP:journals/corr/abs-1803-02999}. 

Formally, given a task $\tau_i$ associated with a dataset sample 
$\mathcal{D}=\{\mathcal{S}, \mathcal{Q}\}$, MAML optimizes the parameters towards the optimal target in Equation \ref{eq:2} via two steps.
In the meta-training step, or inner step for short, the algorithm updates the parameters to $\theta^{'}$ on the support set $\mathcal{S}$. The updated parameters $\theta^{'}$ is computed by  loss $\mathcal{L}^{(0)}$ and gradient descent steps on $\mathcal{S}$ with an update step size $\alpha$, 

\begin{equation}
    \theta'_i = \theta - \alpha \nabla_\theta\mathcal{L}^{(0)}_{\tau_i}(f_\theta)
\end{equation}

The meta-testing step, or meta step for short, takes in the updated model $f(\theta^{'})$ and optimizes the parameters $\theta$ based on the loss $\mathcal{L}^{(1)}$ of  $f(\theta^{'})$ across $\mathcal{Q}$. While the objective is computed using the updated model parameters $\theta^{'}$, the meta-optimization is performed over the
model parameters $\theta$, that the optimal model parameters are,

\begin{align}
    \theta^* 
&= \arg\min_\theta \sum_{\tau_i \sim p(\tau_i)} \mathcal{L}_{\tau_i}^{(1)} (f_{\theta'_i}) \\
&= \arg\min_\theta \sum_{\tau_i \sim p(\tau_i)} \mathcal{L}_{\tau_i}^{(1)} (f_{\theta - \alpha\nabla_\theta \mathcal{L}_{\tau_i}^{(0)}(f_\theta)})
\end{align}

The model parameters $\theta$ are updated with a meta-update step size $\beta$,

\begin{equation}
    \theta \leftarrow \theta - \beta \nabla_{\theta} \sum_{\tau_i \sim p(\tau)} \mathcal{L}_{\tau_i}^{(1)} (f_{\theta - \alpha\nabla_\theta \mathcal{L}_{\tau_i}^{(0)}(f_\theta)})
\end{equation}




\section{Cross-lingual Adaption Model-Agnostic Meta-Learning}

\subsection{Meta Task Formulation}

Cross-lingual Natural Language Understanding aims to transfer knowledge from high resource languages to target low resource languages, that the meta task formulation of MAML needs to adapt to such changes.

To perform cross-lingual adaption, we no longer randomly form meta learning tasks $\tau_i$ from one distribution $\mathbb{D} = \{(x,y)\}$, where $\mathbb{D}$ stands for the available feature vectors $x$ and ground truth labels $y$ pairs. We split $\mathbb{D}$ into $\{\mathbb{D}_{{l_i}}\}_{{l_i} \in \mathbb{L}}$ by language $l$ of language set $\mathbb{L}$, and form the associated dataset $\mathcal{D}$ of meta learning tasks $\tau_i$ as $\mathcal{D} = (\mathcal{S}, \mathcal{Q})$ with the following two steps:

\begin{enumerate}
    \item Sample two subset of languages for support set $\mathcal{S}$ and query set $\mathcal{Q}$: $\{l_i\}_S \subset \mathbb{L}, \{l_i\}_Q \subset \mathbb{L}$. 
    \item Sample support set $\mathcal{S}$ with $K$ data-points and query set $\mathcal{Q}$ with $N$ data points from language subsets of $\mathbb{D}$: \\$\mathcal{S} = \{(x^k,y^k)\}_{k=1}^K \subset \bigcup_{\{l_i\}_S}\mathbb{D}_{l_i} \subset \mathbb{D}$, \\$\mathcal{Q} = \{(x^n,y^n)\}_{n=1}^N \subset \bigcup_{\{l_i\}_Q}\mathbb{D}_{l_i} \subset \mathbb{D}$.
\end{enumerate}

Where the language subset size is fixed as a hyperparameter. Thus the sampled tasks $\tau_i$ can simulate the process of cross-lingual transfer and allow the meta learning perform cross-lingual fast adaption. 

Due to the characteristics of low-resource cross-lingual transfer, we further explore two settings:
\begin{itemize}
    \item The available resource amount for the cross-lingual transfer is relatively low. To take advantage of all the data, we change the sampling strategy from random sampling to covering all the given data.
    \item Parallel data is often beneficial for cross-lingual learning, we explore the effectiveness of parallelism in the inner-step and meta-step.
\end{itemize}

\subsection{Training Procedure}

We show the pipeline of training procedure in Algorithm \ref{alg:XLA-MAML}.
 We first initialize our model $f(\theta)$ with multilingual pretrained Masked Language Models (mMLM) such as mBERT\cite{devlin2019bert}, which jointly trains monolingual masked language models across multiple languages. These models have been proven strong performance on cross-lingual transfer and serves as effective model initialization. 
 In the second step we finetune our model on English monolingual data as a second pretraining step. This step allows the model to take benefits of the high resource data and serves as a baseline model.
 
 We formulate meta-learning tasks as formerly stated. The meta dataset sample $\mathcal{D}$ aims to perform fast adaption under cross-lingual transfer setting.
We perform the inner step and update the model to $f(\theta'_i)$ on the support set $\mathcal{S}$ loss $\mathcal{L}^{(0)}$ for hyperparameter set inner steps by gradient descent.   
We perform the meta step and optimize the parameters $\theta$ based on the loss $\mathcal{L}^{(1)}$ of $f(\theta^{'})$ on the query set $\mathcal{Q}$.

 We perform either zero-shot evaluation or few-shot learning on the target languages. For zero-shot evaluation, we directly evaluate the performance of the models on unseen test set $\{x_{tgt}\}$ of the target languages. For few-shot learning, we use a small development set of the target language to finetune the model as the standard supervised learning, and then evaluate the performance on the test set.

\begin{table}[t]
\centering

\renewcommand\arraystretch{1.3}
\begin{tabular}{lcccc}
\hline
\textbf{Dataset} & \textbf{\#Train} & \textbf{\#Dev} & \textbf{\#Test} & \textbf{Metric} \\ 
\hline
\textbf{MNLI} & 392,702 & 20,000 & 20,000 & Acc.\\
 \textbf{SQuAD} & 87,599 & 34,726 & - & EM/F1\\

\hline
\end{tabular}

\caption{The data statistics of the high resource English datasets. Dev is short for development set. The test set of SQuAD is not public available.
}
\label{tab:monodataset}
\end{table}


\begin{table*}[t]
\centering
\small
\setlength\tabcolsep{3pt}
\renewcommand\arraystretch{1.3}
\begin{tabular}{llcccccccccccccccc}
\hline
&& \multicolumn{15}{c}{\textbf{\#Examples}}\\
\textbf{Dataset} &\textbf{Split} & \textbf{ar} & \textbf{bg} & \textbf{de} & \textbf{el}&\textbf{en}& \textbf{es}&\textbf{fr}&\textbf{hi}&\textbf{ru}&\textbf{sw}&\textbf{th}&\textbf{tr}&\textbf{ur}&\textbf{vi}&\textbf{zh}\\
\hline

\multirow{2}{*}{\textbf{XNLI}}& Dev &2,500&2,500&2,500&2,500&2,500&2,500&2,500&2,500&2,500&2,500&2,500&2,500&2,500&2,500&2,500 \\
 &Test &5,000&5,000&5,000&5,000&5,000&5,000&5,000&5,000&5,000&5,000&5,000&5,000&5,000&5,000&5,000\\
\hline
\multirow{2}{*}{\textbf{MLQA}} &Dev& 517&-& 512&-&1148&500&-&507&-&-&-&-&-&511&504\\
 &Test& 5,335&-& 4,517&-&11,590&5,254&-&4,918&-&-&-&-&-&5,495&5,137\\
\hline

\hline
\end{tabular}

\caption{The statistics of the multilingual datasets by languages. Dev is short for development set. The data of XNLI is parallel across languages generated by human translation from the English Data. The data of MLQA is partially parallel generated by automatic matching of target language Wikipedia articles to English Wikipedia, and then verified by crowd-sourced human annotation. 
}
\label{tab:xldataset}
\end{table*}

\section{Experiments}

\subsection{Datasets}

We show the dataset details in Table \ref{tab:monodataset} and Table \ref{tab:xldataset}.

\subsubsection{Natural Language Inference}

Natural Language Inference (NLI) is the task of determining whether a given pair of \textit{hypothesis} and \textit{premise} is true (entailment), false (contradiction), or undetermined (neutral). It can be considered as a sentence pair classification task.

\paragraph{MNLI} The Multi-Genre Natural Language Inference (MNLI) \cite{N18-1101} dataset has 433k sentence pairs annotated with textual entailment information. The dataset covers a range of genres of spoken and written text and supports cross-genre evaluation. 

In our experiment MNLI is applied as the high resource English dataset in the second pretraining step.

\paragraph{XNLI} The Cross-Lingual Natural Language Inference (XNLI) \cite{conneau2018xnli} dataset is build for evaluation of cross-lingual sentence understanding methods. It consists of a crowd-sourced collection of 2500 development and 5000 test hypothesis-premise pairs for the MultiNLI corpus with their textual entailment labels. The pairs translated into 14 languages: French (fr), Spanish (es), German (de), Greek (el), Bulgarian (bg), Russian (ru), Turkish (tr), Arabic (ar), Vietnamese (vi), Thai (th), Chinese (zh), Hindi (hi), Swahili (sw) and Urdu (ur), which results in 112.5k annotated pairs. 

The dataset supports both zero-shot evaluation and few-shot training for cross-lingual transfer. Under the zero-shot evaluation setting, the model is directly tested on unseen target language with only the high-resource language English data available from MNLI and low-resource auxiliary languages data from the development set of XNLI. Under few-shot learning setting, the model is further finetuned by the development set of target language data of XNLI, and then tested on the target test data.



%

\subsubsection{Question Answering}

Question Answering (QA) is the task of answering a question based on a given support document. While there are various QA task settings including multi-choice QA, cloze style QA and so on, in this paper we focus on reading comprehension dataset, where the answer to every question is a segment of text (a span) from the corresponding reading passage. 

\paragraph{SQuAD} The Stanford Question Answering Dataset (SQuAD) \cite{rajpurkar-etal-2016-squad, rajpurkar2018know} consisting of 100,000+ questions posed by crowd-workers on a set of Wikipedia articles. For each question, three candidate answers spans are given, which are segment slices of the paragraph. The evaluation of model prediction is based on Exact Match (EM), which measures the percentage of predictions that match any one of the ground truth answers exactly, and F1 score, which measures the average overlap between the prediction and ground truth answer.  

In our experiment, SQuAD 1.1 \cite{rajpurkar-etal-2016-squad} is applied as the high resource English dataset in the second pretraining step. There are no un-answerable questions in the training data.

\paragraph{MLQA} Multilingual Question Answering (MLQA) \cite{lewis2020mlqa} is a benchmark dataset for evaluating cross-lingual question answering performance. MLQA consists of over 5K extractive QA instances (12K in English) in SQuAD format in seven languages - English(en), Arabic(ar), German(de), Spanish(es), Hindi(hi), Vietnamese(vi) and Simplified Chinese(zh). The development set of languages are highly but not completely parallel, detailed statistics are shown in Table \ref{tab:xldataset}.

There are other cross-lingual question answering datasets such as Cross-lingual Question Answering Dataset (XQuAD) \cite{artetxe-etal-2020-cross} and 
Information-Seeking Question Answering in Typologically Diverse Languages (TyDiQA) \cite{tydiqa}. XQuAD does not have a development set that it does not support few-shot training scenario. TyDiQA has a relatively small size of English training data, which has a mis-match with our experimental settings of English as a high resource language.

For both NLI and QA, we train with the same model described in \citet{devlin2019bert}, except for the pretrained layers are changed to mMLMs.

\begin{table*}[t]
\centering
\scriptsize
\setlength\tabcolsep{3pt}
\renewcommand\arraystretch{1.2}
\begin{tabular*}{\textwidth}{l@{\hskip32pt}cccccccccccccccc}
\hline
\textbf{Method} & \textbf{ar} & \textbf{bg} & \textbf{de} & \textbf{el}&\textbf{en}& \textbf{es}&\textbf{fr}&\textbf{hi}&\textbf{ru}&\textbf{sw}&\textbf{th}&\textbf{tr}&\textbf{ur}&\textbf{vi}&\textbf{zh}& \textbf{avg}\\
\hline
\multicolumn{17}{l}{\textit{Zero-shot Cross-lingual Transfer}}\\
\hline
mBERT (Ours)& 65.65&68.92&70.82&67.09&81.32&74.65&73.15&60.30&68.64&51.32&55.15&62.81&58.32&70.10&69.04&66.48\\
\textit{One aux. lang.}\\
X-MAML hi $\rightarrow X$ & 66.35 & 70.84 & 73.39 & 69.38 & 82.33 & 76.71 & 75.63 & - & 70.54 & 49.94 & 56.01 & 63.79 & 62.16 & 72.69 & 72.18 & - \\ 
XLA-MAML hi $\rightarrow X$& 67.78 &72.36&\textbf{74.23}&\textbf{70.90}&82.04&76.95&75.77&-&\textbf{71.98}&50.36&60.02&64.87&64.11&73.53&73.43&- \\

\textit{Two aux. lang.}\\
X-MAML best pair & 67.35 & 70.80 & 73.71 & 69.80 & \textbf{82.37} & 76.85 & 75.93 & 64.49 & 71.10 & 51.84 & 57.45 & 64.17 & 62.81 & 72.95 & 72.91 & 68.96 \\
XLA-MAML same pair&67.82&72.04&72.67&69.42&81.44& 76.17& 74.61&64.85& 71.64&53.71&60.02&64.59&64.07&72.71&73.65&69.29\\
XLA-MAML best pair&\textbf{67.91}&\textbf{72.77}&74.00&70.63&82.06&\textbf{77.03}&\textbf{75.96}&\textbf{65.03}&71.66&\textbf{53.82}&\textbf{60.34}&\textbf{66.29}&\textbf{64.31}&\textbf{73.73}&\textbf{74.01}& \textbf{69.97}\\
\hline
\multicolumn{17}{l}{\textit{Few-shot Cross-lingual Transfer}}\\
\hline
mBERT (Ours)& 67.98&72.59&73.79&69.88&81.2&75.15&74.33&65.53&71.48&57.88&63.17&65.83&63.41&71.54&73.09&69.79\\
One aux. lang.\\
X-MAML \textit{sw} $\rightarrow X$ & 68.04 & 71.70 & 73.93 & 70.06 & 82.57 & 77.72 & 75.93 & 64.91 & 72.16 & - & 60.94 & 65.25 & 63.57 & 73.05 & 74.29 &  \\
XLA-MAML \textit{sw}$\rightarrow X$&\textbf{69.32}& 71.78 & \textbf{73.47}&70.37 & 81.90 &  76.40 & 75.65 & 65.64& 71.62 & - & 63.51& \textbf{66.16}& \textbf{63.73} & 71.80& \textbf{74.65} \\

Two aux. lang.\\
X-MAML best pair & 68.09 & 72.02 & \textbf{73.93} & 69.48 & \textbf{82.79} & \textbf{77.99} & 76.69 & 65.49 & \textbf{72.28} & 58.59 & 61.24 & 66.12 & 63.21 & \textbf{73.07} & 73.43 & 70.30 \\
XLA-MAML same pair & 68.66&\textbf{72.75}&73.57&70.35&81.99 &77.23 &\textbf{76.91}&\textbf{65.82}&71.92&\textbf{58.71}& \textbf{63.62}&65.73& 63.70&72.06&74.11& \textbf{70.47}\\
\hline
\multicolumn{17}{l}{\textit{Machine translate}}\\
\hline 
\citet{devlin2019bert} \textit{test}& 70.4 & - & 74.4 & - & 81.4 & 74.9 & - & - & - & - & - & - & 62.1 & - & 70.1 & -\\
\citet{wu-dredze-2019-beto} \textit{train}& 70.8 & 75.4& 74.8& 72.1 & 82.1 & 78.5 & 76.9 & 65.3 & 74.3 & 65.3 & 63.2 & 70.6 & 60.6 & 67.8 & 76.2 & 71.6\\

\hline
\end{tabular*}

\caption{Accuracy results on the XNLI test set for zero-shot and few-shot XLA-MAML. Each columns indicate the target language. `\textsuperscript{*}' denotes the average accuracy excluding the auxiliary languages, which are not comparable to the averages over all languages.
}
\label{tab:nliresults}
\end{table*}
\begin{figure}[t]
\flushleft
\centering

    \includegraphics[width=0.45\textwidth]{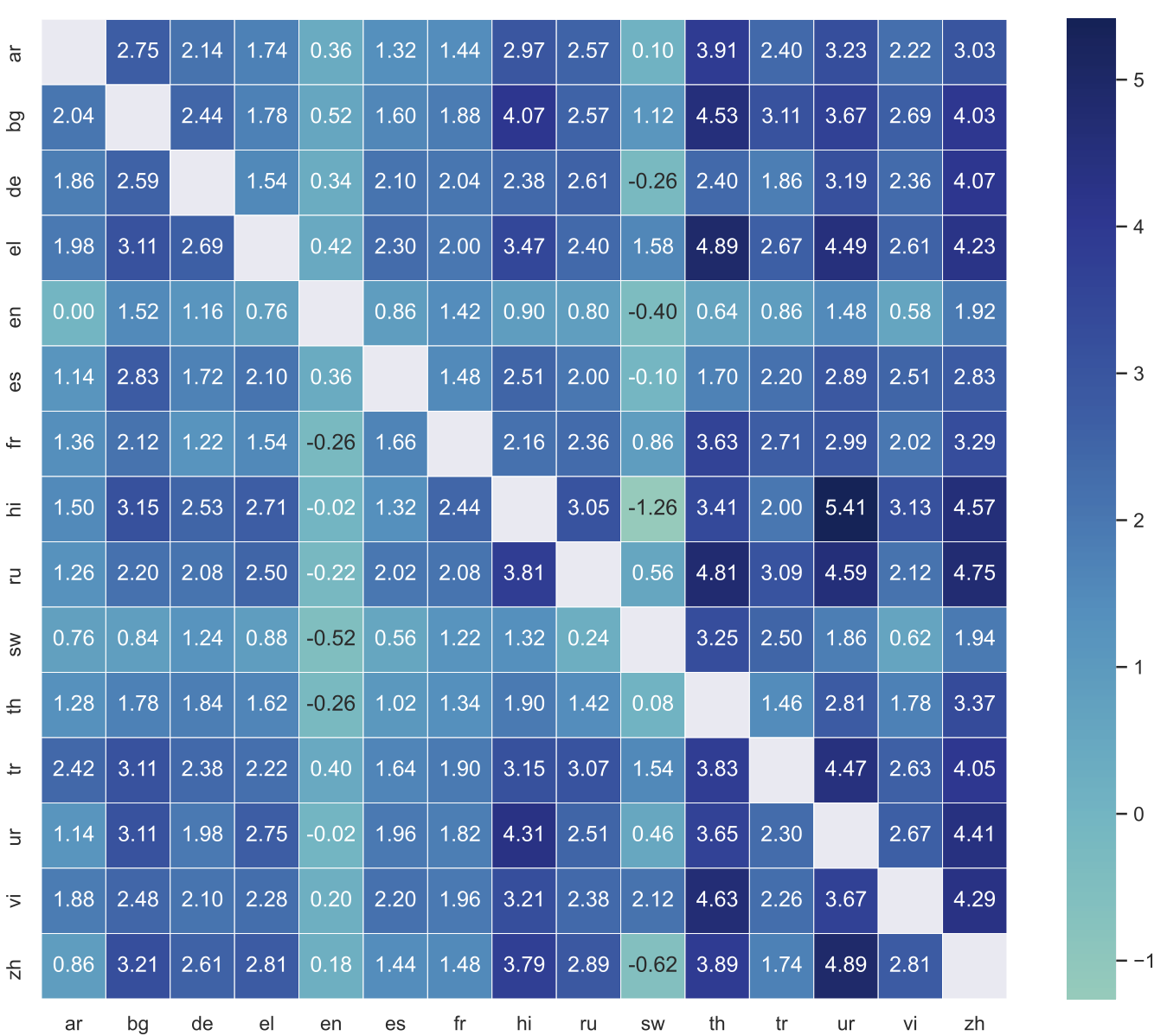}

        \caption{Differences in performance in terms of accuracy scores on the test set for zero-shot XLA-MAML on XNLI using the mBERT model. Rows correspond to auxiliary and columns to target languages. Numbers on the off-diagonal indicate performance differences between XLA-MAML and the baseline model. The coloring scheme indicates the differences in performance (e.g., blue for large improvement).}
    \label{fig:zeronli}
\end{figure}

\subsection{Training setup and hyper-parameters}

We implement our algorithm with the libraries \textit{Pytorch}~\cite{NEURIPS2019_9015}, \textit{Transformers}~\cite{wolf-etal-2020-transformers} and \textit{higher}~\cite{grefenstette2019generalized}. All datasets are loaded with the \textit{Datasets}~\cite{2020HuggingFace-datasets} library.

We use the cache of pretrained models of \textit{Transformers} to initialize the model. 
For MNLI finetuning, we train the model with 32 batch size and 128 max sequence length for 3 epochs. We use AdamW~\cite{DBLP:conf/iclr/LoshchilovH19} optimizer with 2e-5 learning rate. 
For SQuAD finetuning, we train the model with 12 batch size, 384 max sequence length and 128 document stride for 2 epochs. We use AdamW optimizer with 3e-5 learning rate. For both datasets we only use the training data to finetune the model.

For XLA-MAML, for both datasets we use the same data preprocessing parameters as the English model. We use batch size 8 for both inner-step update and meta-step update. We use learning rate of 1e-5 with SGD optimizer for the inner-step update, and use learning rate of 1e-5, weight decay of 0.01 with Adam optimizer~\cite{DBLP:journals/corr/KingmaB14} and a linear learning rate scheduler for the meta-step update. NLI models with memory usage of around 10G use 1 minute for each 100 meta steps trained on GeForce GTX 1080Ti and QA models with memory usage of around 22G use 2 minutes for each 100 meta steps trained on TITAN RTX. 

Since the low resource languages only have development set, to avoid fitting to the test set and select a stopping criterion of the algorithm, we observe the performance of a few runs on unseen language development set to fix a stopping iteration number. Empirically we use 500 meta iterations per meta-step language for NLI and 100 meta iteration per meta-step language for QA. Similarly, we fix the few-shot finetuning epoch to 1 epoch for both NLI and QA for fair comparison.

\begin{table*}[t]
\centering
\scriptsize
\setlength\tabcolsep{3pt}
\renewcommand\arraystretch{1.2}
\begin{tabular}{lccccccc}
\hline
\textbf{Method} & \textbf{ar} & \textbf{de} & \textbf{en} & \textbf{es}&\textbf{hi}& \textbf{vi}&\textbf{zh}\\
\hline
\multicolumn{8}{l}{\textit{Zero-shot Cross-lingual Transfer}}\\
\hline
mBERT (Our baseline)& 48.88/34.73& 66.69/52.42 & 80.41/67.11 &69.13/54.69 & 42.63/29.85 & 55.64/42.04&56.04/42.46\\
One aux. lang.\\
X-MAML \textit{hi} $\rightarrow X$ &
53.99/41.09&
69.15/
55.52&
79.32/
65.97&
70.15/
56.08&
-
&
59.68/
46.84&
61.87/
48.16
\\
XLA-MAML \textit{hi} $\rightarrow X$& 55.05/42.81&69.87/57.10& 79.69/66.63&70.72/57.53& -& 62.28/50.28& 62.82/49.93 \\
\hline
\multicolumn{8}{l}{\textit{Few-shot Cross-lingual Transfer}}\\
\hline
mBERT (Ours)& 56.28/44.48& 72.28/60.02 & 81.08/68.01 & 72.34/58.98& 51.98/40.08 & 62.47/50.06&63.80/50.67\\
One aux. lang.\\
XLA-MAML \textit{hi}$\rightarrow X$&57.96/46.29&72.39/59.57& 80.86/67.64& 72.70/59.34& 58.60/47.05&64.29/52.16&65.49/52.77\\
\hline
\multicolumn{8}{l}{\textit{Machine translate}}\\
\hline
\citet{lewis2020mlqa} \textit{test} & 33.6 / 20.4 & 57.9 / 41.8 & 80.2 / 67.4 & 65.4 / 44.0 & 23.8 / 18.9 & 58.2 / 33.2 & 44.2 / 20.3\\
\citet{lewis2020mlqa} \textit{train} & 51.8 / 33.2 & 62.0 / 47.5 & 77.7 / 65.2 & 53.9 / 37.4 & 55.0 / 40.0 & 62.0 / 43.1 & 61.4 / 39.5 \\
\hline

\end{tabular}

\caption{Accuracy results on the MLQA test set for zero-shot and few-shot XLA-MAML. Each columns indicate the target language.
}
\label{tab:qaresults}
\end{table*}
\subsection{Results}

\paragraph{NLI} We employ zero-shot and few-shot learning for XNLI dataset with XLA-MAML and report the results in Table \ref{tab:nliresults} and Figure \ref{fig:zeronli}. In Table \ref{tab:nliresults}, the scores are reported by 3 run average on XNLI test set. For fair comparison to X-MAML~\citet{nooralahzadeh2020zero}, that they rely on a weaker MNLI pretrained model, we use their public finetuning code and load a MNLI pretrained model trained with same hyperparameters with our baseline, which achieved higher performance than the scores reported in \citet{nooralahzadeh2020zero}.

We compare results of XLA-MAML and X-MAML only using one auxiliary language development set as auxiliary data. 
To simulate fast adaption from high resource \textit{en} to low resource language, we use \textit{en} for the inner-step update and \textit{aux. lang} for the meta-step update. Here the we use the development set of XNLI \textit{en} for inner step which is a subset of MNLI development set. The experimental results show no statistical significant difference in drawing samples from XNLI or MNLI English development set. The results show that our method can achieve comprehensive improvement over the languages. \citet{nooralahzadeh2020zero} reported \textit{hi} to be the most beneficial auxiliary language for their algorithm, XLA-MAML achieves further improvement compared to X-MAML with the sample cross-lingual data resource. To be noted, under our setting \textit{hi} is not the most beneficial auxiliary language. We show the zero-shot evaluation results of all auxiliary language and target language pairs in Figure \ref{fig:zeronli}. We observe the auxiliary languages beneficial for zero-shot scores for up to 5.4\% accuracy for some language pairs such as $ur \rightarrow hi$. Significant improvement was achieved for all auxiliary languages  except for \textit{en}, where is reasonable that no cross-lingual adaption is performed and also \textit{sw}, which matches with the results reported by \citet{nooralahzadeh2020zero}, where X-MAML lowered the scores with \textit{sw} as auxiliary language. We consider this caused by the typological commonalities of \textit{sw}. 

We also report their result of two auxiliary language pairs. We reproduce the best auxiliary language pairs results reported in \citet{nooralahzadeh2020zero}. We show the corresponding results, which the two languages either as inner-step update or meta-step update. We also show the best language pairs results for XLA-MAML.

For few-shot learning learning, we show results of two settings.
In the one auxiliary and two auxiliary setting, since we have a development set on the target language for finetuning, we use \textit{tgt. lang} as the inner-step and \textit{aux. lang} as the meta-step. 
 Under few-shot setting, with the supervision of target language, the effect of auxiliary languages are limited for all MAML methods.
 The search of best language for few-shot learning setting is enormous.
We compare with the best auxiliary language setting of X-MAML and achieve higher or comparable results. 
 Under few-shot setting, with the supervision of target language, the effect of auxiliary languages are limited for all MAML methods.
In addition, we report the results from \citet{devlin2019bert} that use machine translation at test time (TRANSLATE-TEST) and results from \citet{wu-dredze-2019-beto} that use machine translation at training time (TRANSLATE-TRAIN) for reference. In TRANSLATE-TRAIN, 433k translated sentences are used for fine-tuning.

\paragraph{QA}

We employ zero-shot and few-shot learning for MLQA dataset with XLA-MAML and report results in Table \ref{tab:qaresults}.We use similar settings as NLI for QA results. The code is not available for X-MAML QA and the paper did not report results on mBERT, that we reproduce their results with our own implementation.
We report results for one-auxiliary language \textit{hi} 
. Few-shot learning was not investigated by X-MAML in \citet{nooralahzadeh2020zero}, that we show our results comparing to few-shot baseline. Similar to NLI, we report TRANSLATE-TEST and TRANSLATE-TRAIN \citet{lewis2020mlqa} which are strong baselines, where TRANSLATE-TRAIN is finetuned with 100k translated examples.

\section{Analysis}

\subsection{Results on XLM-R}

\citet{conneau-etal-2020-unsupervised} introduced XLM-R, which uses larger models and more data than mBERT to pretrain an mMLM. We report zero-shot evaluation results on XLM-$R_{base}$ to show that our algorithm comprehensively achieves improvement on stronger baselines. We show the results on XNLI in Table \ref{tab:xlm}. As we can see our model out-performs the zero-shot baseline for around 2\% acc. in average, and also exceeds XMAML baseline with hi, demonstrating the generalization ability of our model on stronger mMLM baselines.

\begin{table}[t]
\centering
\footnotesize
\setlength\tabcolsep{2pt}
\renewcommand\arraystretch{1.2}
\begin{tabular}{llccccccc}
\hline
\textbf{Model}&\textbf{aux.} & \textbf{ar}  & \textbf{es}&\textbf{sw}& \textbf{vi}&\textbf{zh} & \textbf{avg}.\\
\hline
XLM-$R_{base}$&- & 71.71   & 78.56  & 65.53 & 74.41 & 72.08 & 73.96\\
XLA-MAML&hi   & 73.63  & 79.54  & 67.15 & 77.36 & 76.24 & 75.81\\
X-MAML& hi  &72.87&79.04&65.07&76.49&76.41&75.34\\
XLA-MAML&bg+de & 73.21   & 79.50  & 66.23 & 76.68 & 76.68 & 75.65\\
\hline
\end{tabular}

\caption{Results on XLM-$R_{base}$ baseline. \textbf{aux.} stands for the auxiliary languages used for training. The avg. score is reported on 15 languages average.
}
\label{tab:xlm}
\end{table}

\subsection{Effects of Sampling Strategy}

\begin{table}[t]
\centering
\footnotesize
\setlength\tabcolsep{2pt}
\renewcommand\arraystretch{1.2}
\begin{tabular}{llccccccc}
\hline
\textbf{Task}&\textbf{Method} & \textbf{ar} & \textbf{de} & \textbf{en} & \textbf{es}&\textbf{hi}& \textbf{vi}&\textbf{zh}\\
\hline
\multirow{4}{*}{NLI}&hi r. & 66.17  & 73.11 & 81.64 & 76.03  & - & 71.87 & 72.91\\
&hi c. & 67.78  & 74.23 & 82.04 & 76.95  & - & 73.53 & 73.43\\
&bg+de r. & 66.90  & - & 81.26 & 75.45  & 62.83 & 72.43 & 72.28\\
&bg+de c. & 67.60  & - & 81.52 & 76.20  & 62.95 & 72.79 & 72.51\\
\hline
\multirow{4}{*}{QA}&hi r. &  53.96 & 68.78 & 80.17 & 70.41 & - & 60.66& 62.69\\
&hi c. & 53.99  & 69.87 & 79.69 & 70.72  & - & 62.28 & 62.82\\
&ar+hi r. & -  & 71.33 & 80.18 & 71.87  & - & 62.77 & 63.57\\
&ar+hi c. & -  & 71.37 & 80.27 & 71.94 & -  & 63.45 & 63.72\\
\hline

\end{tabular}

\caption{Results of different sampling strategies for meta learning. The language codes in `Method' denotes the languages used for XLA-MAML. `r.' denotes random sampling. `c.' denotes covering all examples.
}
\label{tab:sampling}
\end{table}

We report the results for sampling strategy in Table \ref{tab:sampling}. Since the development set for XLA-MAML is relatively small, unlike in few-shot learning the data for MAML is larger, we consider it important to cover all of the development data during the meta learning update. We use compare two sampling strategies, the first strategy randomly samples a batch of examples at each meta learning episode, the second strategy preprocesses the batches that they cover all of the examples. We show results on one auxiliary language and one auxiliary language bi-pair. There is a significant performance drop when the samples are randomly drawn at each episode, reflects the importance of making full use of all data for low-resource transfer learning. 

\subsection{Effects of Parallel Data}

\begin{table}[t]
\centering
\footnotesize
\setlength\tabcolsep{2pt}
\renewcommand\arraystretch{1.2}
\begin{tabular}{lccccccc}
\hline
\textbf{Method} & \textbf{ar} & \textbf{de} & \textbf{en} & \textbf{es}&\textbf{hi}& \textbf{vi}&\textbf{zh}\\
\hline
hi non. & 67.22  & 72.91 & 82.00 & 76.67  & - & 72.79 & 72.08\\
hi p. & 67.78  & 74.23 & 82.04 & 76.95  & - & 73.53 & 73.43\\
bg+de non. & 66.97  & - & 81.14 & 75.95  & 63.15 & 72.63 & 73.21\\
bg+de p. & 67.60  & - & 81.52 & 76.21  & 62.95 & 72.79 & 72.51\\
\hline

\end{tabular}

\caption{Results of parallelism for meta learning. The language codes in `Method' stands for the languages used for XLA-MAML. `non.' stands for non-parallel training. `p.' stands for parallel training.
}
\label{tab:parallel}
\end{table}

Parallel data is reported useful for cross-lingual representation alignment~\cite{DBLP:conf/iclr/LampleCDR18} and mMLM pretraining~\cite{lample2019cross}. The XNLI dataset is completely parallel, that it is possible for us to investigate the effectiveness of parallelism of inner-step and meta-step. The results on one auxiliary language and one auxiliary language bi-pair are shown in Table \ref{tab:parallel}. We can see that parallelism gains all-round but margin improvement. Taking account that the parallel data is relatively tiny, we consider parallelism useful for XLA-MAML training.


\section{Related Work}

\subsection{Cross-lingual Natural Language Understanding}

Research of cross-lingual natural language understanding has recently been investigated with the release of datasets and benchmarks~\cite{liang-etal-2020-xglue, hu2020xtreme} across a wide range of tasks. Alignment by translating the training data to the target language~\cite{wu-dredze-2019-beto} or the testing data to the source language~\cite{devlin2019bert} stands for strong baselines. Cross-lingual approaches rely on sharing a encoder layer for multilingual languages, by aligning representations to a shared vector space~\cite{DBLP:conf/iclr/LampleCDR18, xie18emnlp, DBLP:journals/tacl/ArtetxeS19} or implicitly jointly multiple language training of monolingual Masked Language Models~\cite{devlin2019bert,lample2019cross,conneau-etal-2020-unsupervised}, with or without multilingual parallel data.

\subsection{Meta Learning for Natural Language Processing}

Meta Learning has demonstrated effectiveness for both monolingual NLP and cross-lingual NLP. Previous studies applied meta Learning to few-shot learning and cross-domain transfer learning. \citet{han-etal-2018-fewrel} investigated meta Learning on the task of few shot relation extraction, \citet{xie18emnlp} studied meta learning for few-shot intent classification. \citet{dou-etal-2019-investigating} validated Meta Learning methods on the GLUE benchmark. \citet{larson2019evaluation} and \citet{yu2018diverse} applied cross-domain transfer learning for classification with meta learning, where each domain is considered as a task.

For cross-lingual NLP, \citet{gu-etal-2018-meta} demonstrated the effectiveness of Meta Learning for Neural Machine Translation over strong baselines of cross-lingual transfer learning.  By considering language pairs as tasks, MAML has shown strong results with as few as 600 parallel sentences. 
\citet{nooralahzadeh2020zero} proposed X-MAML, which is an MAML based method with auxiliary languages data as tasks, which uses the data of auxiliary languages in both the inner-step and meta-step. While they achieved improvement over mMLM baselines, their meta learning procedure cannot perform cross-lingual adaption. Our experimental results show how our method gains superiority over X-MAML.

\section{Conclusion}
In this paper, we propose XLA-MAML, a cross-lingual adaption-based meta learning method using auxiliary languages to improve cross-lingual transfer. The algorithm directly performs cross-lingual adaption by using different languages in the support set and the query set during the meta learning procedure, which encourages the meta-learning algorithm to capture cross-lingual shared knowledge among the auxiliary language and source language, and further improve target language performance.
The experiments on cross-lingual NLI and QA on mBERT and XLM-R show the effectiveness of our method, especially under zero-shot setting. Under few-shot setting, while the supervision of the target language weakens the effect of auxiliary languages, our method still achieves comparable or better results. We also investigate various XL-specific settings for meta learning, including the sampling strategy and the usage of parallelism.



\bibliography{anthology}
\bibliographystyle{acl_natbib}

\appendix

\section{Experimental Details}

For NLI, we follow the standard sentence pair classification input method for pretrained MLMs \cite{devlin2019bert}. We add a special separator token \texttt{[SEP]} to the end of each sentence sequence and concatenate them together. A special classification token \texttt{[CLS]} is added in front of every input example. The token sequence is then encoded by the mMLM. At the output, the \texttt{[CLS]} representation is fed into an output layer for classification.
inputs and outputs into the pretrained model and finetune all the parameters end-to-end. 

For QA, we formulate the task into a sequence tagging task~\cite{devlin2019bert}. For encoding, we represent the input question and passage as a single packed sequence. Similar to NLI, we add a special separator token \texttt{[SEP]} to the end of the question and the support passage and then concatenate them together. A unused special classification token \texttt{[CLS]} is added in front of every input example to align the input with the MLM training phrase. For finetuning, we introduce a start vector $S \min \mathbb{R}^H$ and an end vector $E \min \mathbb{R}^H$. The probability of a token to be the start token is calculated by the softmax of a dot product between $S$ and the token representation $T_i$: $P(i = {start}) = \frac{e^{S*T_i}}{\sum e^{S*T_i}}$ and the end token probability $P(j = {end})$ vice versa. The final score of a candidate span $(i,j)$ is defined as $T_i*S + T_j*E$, where $j \geq i$. The training objective is the sum of the log-likelihoods of the correct start and end positions. During evaluation, if the best prediction is an invalid answer where $j < i$, we search for the k-best prediction until the prediction is valid.



\end{document}